\documentclass[letterpaper]{article}

\usepackage{aaai}
\usepackage{times}
\usepackage{helvet}
\usepackage{courier}
\usepackage{graphicx}
\usepackage{subfigure}
\usepackage{epsfig}
\usepackage{epstopdf}
\usepackage{alltt}
\usepackage[table]{xcolor}
\usepackage{float}

\usepackage{amssymb}
\usepackage{url}

\title{Active Player Modelling}
\author{Julian Togelius$^1$, Noor Shaker$^1$ and Georgios N. Yannakakis$^2$\\$^1$Center of Computer Game Research, IT University of Copenhagen, Copenhagen, Denmark\\$^1$Institute for Digital Games, University of Malta, Malta\\ julian@togelius.com, nosh@itu.dk, georgios.yannakakis@um.edu.mt}

\begin{document}
\maketitle

\begin{abstract}

We argue for the use of active learning methods for player modelling. In active learning, the learning algorithm chooses where to sample the search space so as to optimise learning progress. We hypothesise that player modelling based on active learning could result in vastly more efficient learning, but will require big changes in how data is collected. Some example active player modelling scenarios are described. A particular form of active learning is also equivalent to an influential formalisation of (human and machine) curiosity, and games with active learning could therefore be seen as being curious about the player. We further hypothesise that this form of curiosity is symmetric, and therefore that games that explore their players based on the principles of active learning will turn out to select game configurations that are interesting to the player that is being explored.
 
\end{abstract}

\section{Player modelling}

The practice of trying to understand players through analysing their interactions with the games they play is becoming more and more common and important. Under the labels ``player modelling'' and ``game data mining'', researchers from various fields (game studies, game design, computational intelligence and statistical machine learning) are bringing learning algorithms and statistical techniques to bear on the logs of (usually multiple, sometimes huge numbers of) players playing games. What aspects of the interaction the game logs, what features are used for the modelling, and what the object of the modelling is varies widely; this contributes to a considerable diversity among approaches to player modelling.

This is not the place for an overview or taxonomy of player modelling; that has been done better elsewhere~\cite{yannakakis13player,smith11an,yannakakis11experience}. However, for the sake of the argument it is important to give a few examples that showcase the types of player modelling research that has been done. To begin with, there is research where unsupervised learning is used to find structure in the space of players. For example,~\cite{drachen09player} used self-organising maps to cluster players of \emph{Tomb Raider: Underworld} into four broad classes, based on data collected on the game developer's servers via telemetry. In this dataset, each individual playthrough is an instance, with time spent in each location, reward collected, fights fought etc. being features. The resulting player classes mostly corresponded with the conceptualisation of players by the designers of the game, but also suggest that some players (the ``pacifists'') played the game in a way that had the designers had not anticipated.

For the purposes of this article, however, we are more interested in supervised learning. In supervised learning, instances have both features and target values, and the task is the train a model that correctly predicts the target value of an unseen instance. The perhaps most straightforward use of supervised techniques for player modelling is to predict some aspect of in-game behaviour based on some other in-game behaviour. Using the same data set of tens of thousands of Tomb Raider player as discussed above,~\cite{mahlmann10predicting} trained decision trees to predict at what stage of the game a player would give up, and in those cases the player would finish the game, how long time the playthrough would take. Such models could be useful in trying to adapt the game so as not to lose players' interest.

But supervised learning could also be used to predict player behaviour or experience outside of the game. In~\cite{shaker10towards}, the authors collected data from hundreds of players playing a clone of \emph{Super Mario Bros}, with procedurally generated levels. The levels were generated using a parameterised level generator, where parameters corresponded to such properties as amounts of enemies and distribution of gaps. After each pair of levels, players where asked which of the levels they thought was most fun, or challenging or most frustrating. Further, numerous aspects of player behaviour (time spent running, number of jumps etc) was collected. This yielded a dataset where each instance is a playthrough, the features includes both level generator parameters and player behaviour, and the class label is based on the player's opinion about the level. When training neural networks on this dataset, it was found that challenge, fun and frustration could be predicted with high accuracy. This is a useful result in itself, as it provides a way of judging the quality of levels in relation to specific players, but the trained models can also be used to automatically create levels that are as fun/challenging/frustrating as possible for specific player. This is done through keeping those features which describe player behaviour constant, and searching for such combinations of level parameters as maximise the predicted player experience.

\subsection{Data scarcity}

It might seem that there is a lot of potential in these techniques, and indeed there is. But there is also a problem: data is scarce, and good data is very scarce. This might seem an odd thing to say in the ``age of data'' where seemingly every device is online and phoning home to deliver data about how it has been used. The scarce resource is really players and their attention. Collecting data from just a few hundred players for the experiments described above was a nontrivial effort, requiring the extensive use of the authors' social networks. Getting people to play a free game online is surprisingly hard, because there are so many other free games available nowadays.

The situation is different when collecting data from a commercial game, which was developed for playability rather than for research purposes. The Tomb Raider: Underworld dataset used in the first example above contained millions of play sessions, as the game (like many, perhaps most, current console games) phoned home with metrics each time a player finished a level. However, this data does not include any data on player experience, or indeed any data external to the game itself. Much player modelling research is dependent on relating in-game behaviour to external data sources. If we would have wanted to administer questionnaires related to the playing experience, connect electrodes to the players to measure skin resistance, extract data from players' Facebook profiles or do any similar data collection, this would have meant an extra effort which would have severely limited the number of players from which data could have been gathered.

Further, the masses of player data that can be collected from commercial games derives from the game configuration that the players actually played. If the purpose of the player modelling is to investigate how players respond to different configurations of the game (different levels, tweaked character capabilities, changes in the game rules etc), representative variations of the games need to be tested with players. This is most likely to be expensive, and severely limit the players you can collect information from (e.g. players who opt-in to an ``experimental feature evaluation'' program or similar).

At this point, it should be clear that data scarcity is a serious issue. But while not much discussed in player modelling and game data mining research, it is an issue that is much discussed within machine learning research. In particular, a very interesting remedy has been suggested and used to good effect in other application domains, namely \emph{active learning}.


\section{Active learning}

The core idea of active learning is that a learning algorithm learns better and faster if it is allowed to choose for itself which examples to learn from.

In standard supervised learning, a set of labelled instances (tuples) are considered as given, and the algorithm is free to learn from this set in any order. In active learning, there is assumed to be a large (potentially infinite) set of unlabelled instances, but a limit on how many instances can be labelled (alternatively, a cost associated with labelling the instances). The task of the active learning algorithm is therefore to select which instances to label. Every time an active learning has learned something new, it looks at what its current best models are building on those instances that have already been labelled, and selects a new instance for labelling so as to learn as best as possible with a limited number of labels.

Which instances does an effective active learning choose to explore (label)? Intuitively, it chooses the next instance so that it will learn as much as possible from it. Technically, there are several different selection strategies that give different results and might be more or less easy to implement into particular types of supervised learning algorithm. For example, for probabilistic learning algorithms that model their own uncertainty, the next sample point could be where the current model is most uncertain. Alternatively, for learning algorithms which are based on committees or ensembles (i.e. where more than one model is learned), the algorithm might choose to sample where the models disagree most (the \emph{query by committee} selection strategy). There are also other ways to calculate the expected maximum improvement of the model. Several selection strategies are discussed in a recent survey of active learning techniques~\cite{settles09active}, which also explains some of the theory behind active modelling and some examples of its empirical success in application domains such as speech recognition and information extraction.

One method that stands out as potentially useful here is that of~\cite{bongard05nonlinear}, who do a form of active learning using evolutionary computation. The mechanism here is competitive coevolution between models, whose fitness depends on the best available test, and tests, whose fitness depends on their capacity to induce disagreement between the models. This can be seen as an implementation of the query by committee selection strategy, and is readily adaptable to player modelling approaches based on evolutionary computation. 

\section{Active player modelling}

How can we bring the power active learning to bear on player modelling, and thus mitigate the data scarcity problem? First, we need to define what an unlabelled and a labelled instance is. Here, we will regard an unlabelled instance as a potential playthrough, i.e. one which has not yet occurred. A labelled instance is data from an actual playthrough, complete with any external data that might serve as label, e.g. questionnaire data, physiological data or Facebook profile data. For the algorithm to choose which instance to label, means that it chooses a configuration of game and player, lets the selected player play the game, and adds all the collected information as a new labelled instance to its active dataset.

We could make this more concrete by discussing how the player modelling examples we discussed above could be made to fit into an active learning framework. Let us start with the Super Mario Bros example. Here, the space of unlabelled instances is the space of level parameters. Labelling an instance means choosing a level configuration (a set of parameters that can be used to generate a level) and queueing it for play in the app which is used for data collection, so that next time a player plays a pair of levels one of them is the new set of parameters. As the original modelling approach is based on neuroevolutionary preference learning, we could augment it to coevolve player experience models and level parameters. The fitness function for the level parameter population could be that it induces maximum disagreement among the different models that are top-ranked in the model population, or it could be the degree to which the current best model predicts inconsistent or extreme player experiences for that set of level parameters.

In the Tomb Raider example, we could imagine trying to correlate in-game performance to some external data source such as a questionnaire. The space of unlabelled instances here would be defined by players, e.g. their demographics, unless we could also change some aspect of the game in which case we might want to include also the game configuration in that search space. We could use a committee of neural networks to predict e.g. player enjoyment based on demographics and in-game behaviour. For each instance, the active learning algorithm would then select a particular player demographic (e.g. 15-19 years old, inner city, single, female) for which the different neural networks that form part of the committee disagree maximally. A player from this demographic would be selected, and presented with a questionnaire relating to his/her playing experience. Integrating the questionnaire answers with the trace of that player's playing session would yield another labelled instance, which the algorithm would use to update its models before selection where to explore next.

As should be obvious from these examples, the workflow of active player modelling is radically different from the standard approach of first collecting data and then learning from it. This is a necessary complication that has to be tackled in order to enjoy the more efficiency benefits of active learning. But it could also be seen as an opportunity, where the selection mechanism of active learning is turned into a strategy for game adaptation or even a game mechanic.

\section{Curiosity}

The last paragraph of the previous section made you curious. You had expected to read something about how the potential adversary effects of the data selection in active learning could be mitigated, so suddenly reading about how it could be used for adaptation or as a mechanic was not what you had expected. On the other hand, you read the abstract before starting to read the paper, so you vaguely remembered reading something about ``selecting game configurations that are interesting to the player'' in the abstract. In other words you were not completely unprepared for that sentence. Because you have some idea game adaptation and game mechanics, you could also understand some of it. It's not as if you had read some total gibberish such as ``pannkakan hoppar j\"{a}mfota'', which you would not consider it worth trying to understand.

At least, this is why you were curious about that sentence according to Juergen Schmidhuber's theory of curiosity. According to that theory, a optimally curious agent chooses to explore things that are the most interesting. Those are the things which it can currently learn the most about, which in general are those things that are not completely predictable but not completely unpredictable either. This idea was originally articulated in the context of autonomous reinforcement learning agents, which were rewarded for selecting actions whose results improved their world model~\cite{schmidhuber91curious}, and was later developed into a broad framework with applications to developmental psychology and robotics, computational creativity and other fields~\cite{schmidhuber06developmental}. It is worth pointing out that interestingness in this framework is relative to the observer, and that the observer will change as a result of curious exploration. For example, music that is interesting to you is music that is not completely predictable (that annoying hit that plays on the radio all the time) but also not completely unpredictable (that abstract piece of art music that requires an advanced composition degree to make sense of); however, once upon a time the person that you were would most likely have loved that annoying hit song on the radio. In the same way, a curious computational agent will seek out more and more complicated cases as it improves its model.

The idea of active learning is strongly related to Schmidhuber's concept of curiosity, and under many circumstances an active learning system can be seen as an implementation of this concept of curiosity (there are some differences in the details; in particular, Schmidhuber suggests explicitly modelling the expected improvement in the core model using a second model). Therefore, if we equip a game with facilities for effectively modelling its players using active learning, we could see this as the game being curious about its players. Let us see where this perspective takes us.

\section{Curious games playing their players}

When we think of curiosity and games, we usually think of curiosity as one of the main drivers of humans playing games -- this is implicated by several theories of player experience, such as Malone's~\cite{malone81what}. A player that is curious about the game will want to keep playing. This, in turn, square very well with the view that the player learns to play the game while playing, and that a large amount of the fun in playing a game is to be had from the learning process~\cite{koster05a}; the curious player will select game experiences that maximise their potential for learning, and a well-designed game will afford such choices.

Turning the tables and imagining that the game is curious about the player might seem odd at first glance, and not necessarily contributing to player curiosity or player satisfaction. However, we believe that there is an important symmetry here which points to that curious game may contribute to curious players. For a sufficiently good model, the certainty the model has about the player's experience or behaviour should correlate with the certainty the player has about their experience or behaviour. So if the learning algorithm chooses to explore the game configuration where it thinks it can learn the most about how the player behaves, this is likely to be a game configuration where the player can learn much about their own experience. So a sequence of configurations that promotes model learning is likely to be one that promotes player learning.

If this hypothesis is true, active player modelling could be used a form of adaptation system, serving up new game configurations that are likely to maximally interest the player. The selection of new game element so as to maximise learning about the player might even be turned into a game mechanic, in a game where the objective was to collaborate with or outsmart the virtual game master. The extent to which this hypothesis is true is an empirical question which we are committed to investigating. The more elementary hypothesis that active modelling can speed up player modelling -- allow to us to learn good models from fewer game sessions -- is almost certainly true, but still very much worth investigating.


\bibliography{pcg}
\bibliographystyle{aaai}
\end{document}